# A Multi-Stage Hybrid Framework for Automated Interpretation of Multi-View Engineering Drawings Using Vision Language Model

Muhammad Tayyab Khan[1, 3*#], Zane Yong[2#], Lequn Chen[2], Wenhe Feng[1], Nicholas Yew Jin Tan[1], Seung Ki Moon[3]

[1] Singapore Institute of Manufacturing Technology (SIMTech), Agency for Science, Technology and Research (A*STAR), 5 CleanTech Loop, #01-01 CleanTech Two Block B, Singapore 636732, Republic of Singapore

[2] Advanced Remanufacturing and Technology Centre (ARTC), Agency for Science, Technology and Research (A*STAR), 3 CleanTech Loop, #01-01 CleanTech Two, Singapore 637143, Republic of Singapore

[3] School of Mechanical and Aerospace Engineering, Nanyang Technological University, 639798, Singapore

[*]KHAN0022@e.ntu.edu.sg

[#]Equal Contribution

**Abstract.** Engineering drawings are fundamental to manufacturing communication, serving as the primary medium for conveying design intent, tolerances, and production details. However, interpreting complex multi-view drawings with dense annotations remains challenging using manual methods, generic optical character recognition (OCR) systems, or traditional deep learning approaches, due to varied layouts, orientations, and mixed symbolic–textual content. To address these challenges, this paper proposes a three-stage hybrid framework for the automated interpretation of 2D multi-view engineering drawings using modern detection and vision language models (VLMs). In the first stage, YOLOv11-det performs layout segmentation to localize key regions such as *views*, *title blocks*, and *notes*. The second stage uses YOLOv11-obb for orientation-aware, fine-grained detection of annotations, including *measures*, *GD&T* symbols, and *surface roughness* indicators. The third stage employs two Donut-based, OCR-free VLMs for semantic content parsing: the *Alphabetical* VLM extracts textual and categorical information from *title blocks* and *notes*, while the *Numerical* VLM interprets quantitative data such as *measures*, *GD&T* frames, and *surface roughness*. Two specialized datasets were developed to ensure robustness and generalization: 1,000 drawings for layout detection and 1,406 for annotation-level training. The *Alphabetical* VLM achieved an overall F1 score of 0.672, while the *Numerical* VLM reached 0.963, demonstrating strong performance in textual and quantitative interpretation, respectively. The unified JSON output enables seamless integration with CAD and manufacturing databases, providing a scalable solution for intelligent engineering drawing analysis.

## 1 Introduction

Engineering drawings remain central to industrial communication, containing key information that influences manufacturability, product quality, and cost [1–3]. Converting this information into structured, machine-readable formats is critical for enabling automation in computer-aided design (CAD), process planning, and inspection. However, in industrial practice, this extraction process remains largely manual, creating a major bottleneck in digital manufacturing workflows. Manual transcription is time-consuming and non-scalable, particularly for complex drawings where hundreds of annotations are distributed across multiple drawing views. Generic optical character recognition (OCR) tools worsen these issues, as they struggle to interpret rotated text, dense layouts, and symbolic constructs such as geometric dimensioning and tolerancing (GD&T) feature frames [4]. These limitations highlight the need for an OCR-free, layout-aware approach to reliable information extraction.

Deep learning has recently been applied to automate selected aspects of this problem. For instance, Reddy et al. [5] fine-tuned detection models to recognize GD&T symbols in mechanical drawings, while Yazed et al. [6] proposed a hybrid YOLO–OCR pipeline for GD&T recognition. Faltin et al. [7] benchmarked multiple object detectors on symbol detection tasks. However, these studies focus only on isolated aspects of drawing information and fail to generalize to complex, densely annotated drawings or achieve accurate, context-aware parsing across diverse annotation types.

In our prior research [4,8,9], open-source vision language models (VLMs), such as Document Understanding Transformer (Donut) and Florence-2, were fine-tuned on engineering drawing data to extract information from 2D inputs. Additionally, closed-source models such as GPT-4o and Claude-3.5-Sonnet were evaluated for comparative analysis. The fine-tuned VLMs performed well on single part drawings, effectively parsing both symbolic and textual elements. However, their performance declined on complex multi-view drawings due to dense annotations and overlapping views. Processing entire drawings in a single pass caused region overlap, degraded localization, and inconsistent extraction. These limitations motivated the need for a region-aware segmentation strategy to isolate semantically coherent regions before interpretation.

To address these challenges, this work proposes a hierarchical, multi-stage hybrid framework that decomposes drawing interpretation into layout segmentation, orientation-aware annotation localization, and task-specialized semantic parsing. While the framework leverages existing detection and VLMs, its novelty lies in the task-driven orchestration of these components to reflect the semantic structure of engineering drawings. Unlike prior single-stage or end-to-end approaches, the proposed design explicitly isolates views before fine-grained analysis, integrates oriented bounding box (OBB) detection as an intermediate reasoning layer, and employs specialized VLMs for textual and numerical content, enabling robust interpretation of complex multi-view layouts. The framework is developed using two curated datasets: 1,000 drawings for view-level segmentation and 1,406 for annotation-level training and benchmarking. A unified evaluation pipeline ensures consistent performance measurement across stages. The proposed method demonstrates robust generalization to complex, multi-view drawings and supports continual learning to adapt to organization-specific standards.

The remainder of this paper is organized as follows. Section 2 outlines the proposed methodology. Section 3 describes the dataset development process, including layout detection, annotation localization, and VLM training. Section 4 presents the method implementation and results. Section 5 discusses the findings, and Section 6 concludes with the limitations and future directions.

## 2 Methodology Overview

The proposed framework employs a three-stage approach for the automated extraction of structured information from 2D engineering drawings, as shown in Fig. 1. Each stage operates at a distinct level of detail, addressing layout detection, fine-grained annotation localization, and semantic parsing. In the first stage, a region detector (YOLOv11-det) identifies the principal layout components (*Views*, *Title Block*, and *Notes*) and outlines their bounding regions. This segmentation simplifies downstream processing by establishing a clear layout hierarchy and defining the relevant regions of interest. The second stage performs detailed view analysis using orientation-aware object detection (YOLOv11-obb) to localize key annotations such as *GD&T*, *Measures*, and *Surface Roughness* within each detected *view*. The final stage applies two specialized VLMs: an *Alphabetical* model for textual regions (*Title Block* and *Notes*) and a *Numerical* model for quantitative annotations. The outputs from all stages are integrated into a unified JSON representation containing both geometric and textual information, making it directly applicable to downstream manufacturing tasks.

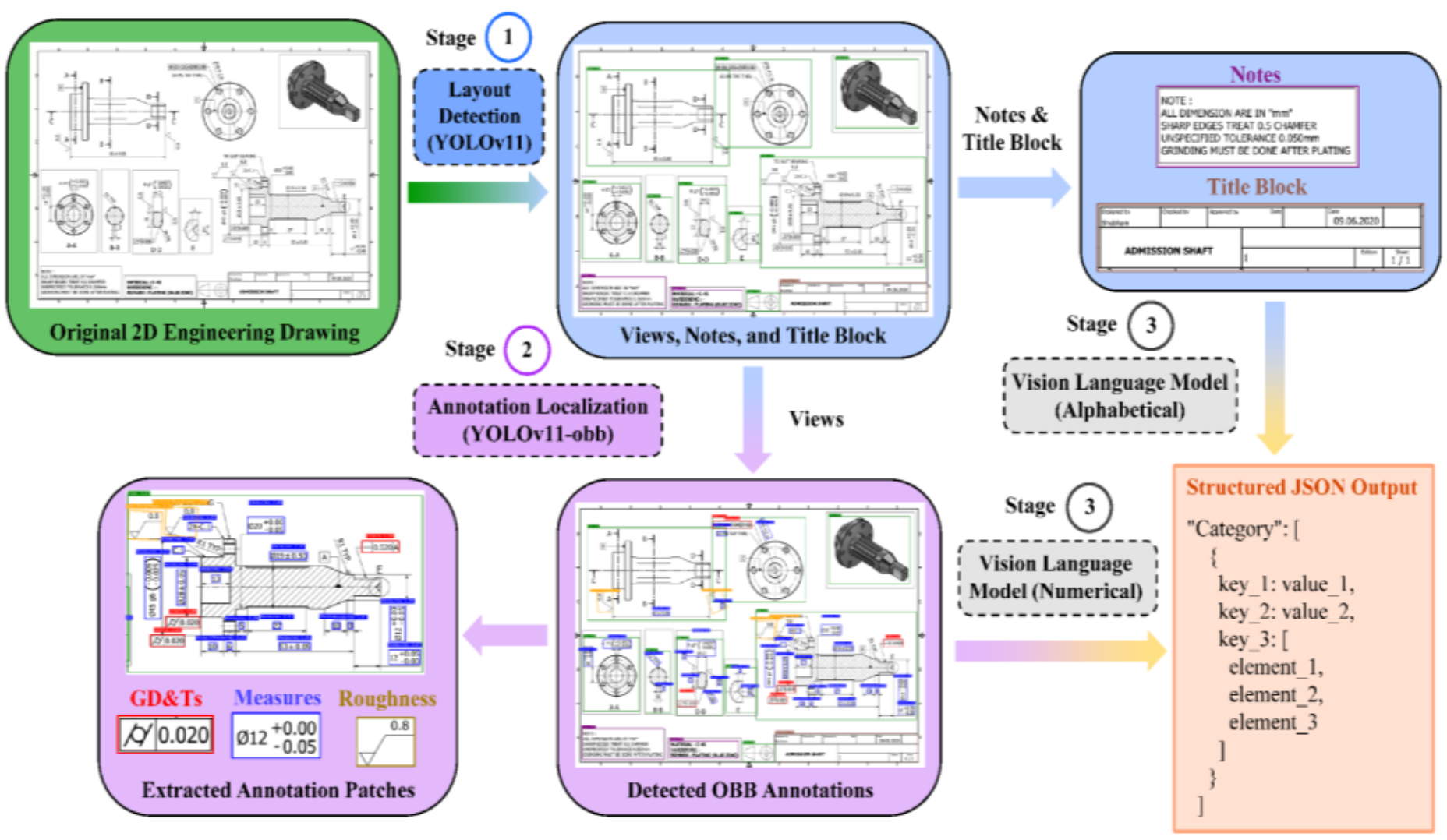

**Fig. 1.** Overview of the proposed three-stage pipeline for extracting structured information from 2D drawings. The framework integrates layout detection, annotation localization, and vision language parsing to generate a unified JSON output for downstream manufacturing tasks.

# 3 Dataset Development and Statistics

The proposed framework relies on domain-specific annotated datasets tailored to both detection and parsing tasks. Separate datasets were curated for each stage to ensure appropriate supervision, class diversity, and balanced evaluation.

## 3.1 Layout Detection Dataset

We collected 1,000 2D mechanical drawings (PDFs and image formats) representing diverse drawing layouts and content types. The dataset encompasses a broad spectrum of engineering documentation, ranging from complex assembly drawings with multiple orthographic and sectional views to simple part drawings. Each drawing was manually annotated with bounding boxes for three region types: *Views*, *Title Block*, and *Notes*. The total label count comprised 3,498 *Views*, 458 *Title Blocks*, and 1,127 *Notes*, indicating that most drawings contain multiple views and occasionally multiple notes. This dataset was used to train the YOLOv11-det model for the first stage. Fig. 2 illustrates the annotated label distribution for YOLOv11-det training. Views constitute the majority of labeled regions, followed by title blocks and notes, consistent with real world engineering drawings where visual content is most prevalent.

## 3.2 Annotation Localization Dataset

For the fine-grained annotation dataset, we compiled a larger set of 1,406 drawings focusing on detailed callouts. These were sourced from online repositories and internal archives to ensure a wide variety of drafting and annotation styles. We annotated three categories within the view regions: *Measures* (including *radii* and *thread* callouts), *GD&T* frames, and *Surface Roughness* symbols. In total, 9,663 *Measures*, 3,215 *GD&Ts*, and 152 *Roughness* instances were annotated. The class distribution, visualized in Fig. 2, is notably imbalanced: *Measures* dominate, *GD&T* frames are fewer, and *Surface Roughness* symbols are relatively rare, which aligns with real-world drawings that typically contain many more dimensional annotations than other types. This dataset was used to train the YOLOv11-obb model employed in the second stage.

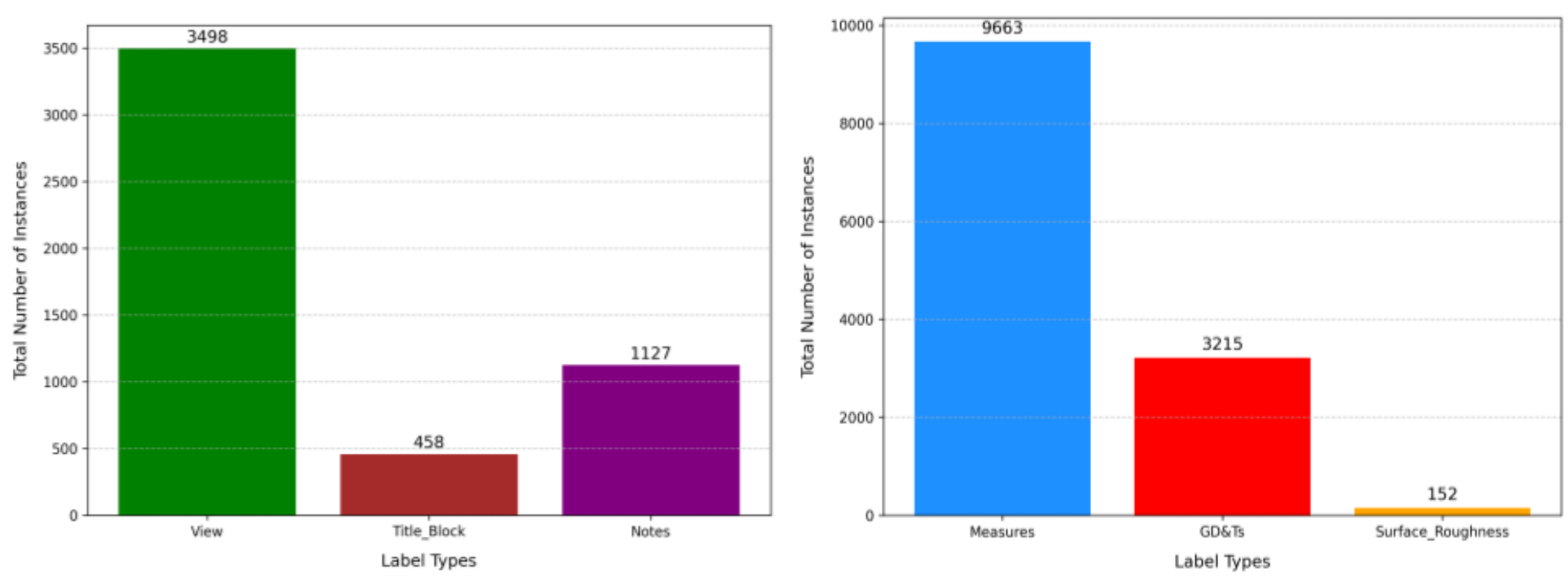

**Fig. 2.** Annotated label distribution for layout region types (*left*) and fine-grained annotation categories (*right*).

### 3.3 VLM Training Dataset

To train and evaluate the two distinct VLMs, we constructed a reference dataset that pairs each detected region or annotation with its ground truth textual content. For every *Title Block* and *Notes* region, structured text and JSON fields were prepared to represent the true information extracted from the drawing. Similarly, for each *GD&T* symbol, *Measure*, and *Surface Roughness* image patch, we recorded the correct numeric or symbolic specification (e.g., tolerance values or roughness parameters). In total, the dataset contained around 15,000 image–text pairs, derived from all annotated regions in the detection datasets. Among these, about 1,585 samples correspond to *Title Block* and *Notes* regions, while around 13,000 samples represent numeric annotations. This structured dataset provides the foundation for subsequent VLM fine-tuning and evaluation, ensuring supervision with precise JSON outputs which supports accurate field extraction and structured information generation. All annotations were carefully validated for quality, and the class proportions were maintained to reflect real-world distributions.

## 4 Detailed Method Implementation and Results

Following dataset preparation, each stage of the proposed pipeline was trained and evaluated using independent supervision to ensure modular reliability and validation. Fig. 3 illustrates a representative example of the complete framework applied to a single engineering drawing, showing the transition from layout detection to annotation localization and vision language parsing, leading to a unified structured JSON output.

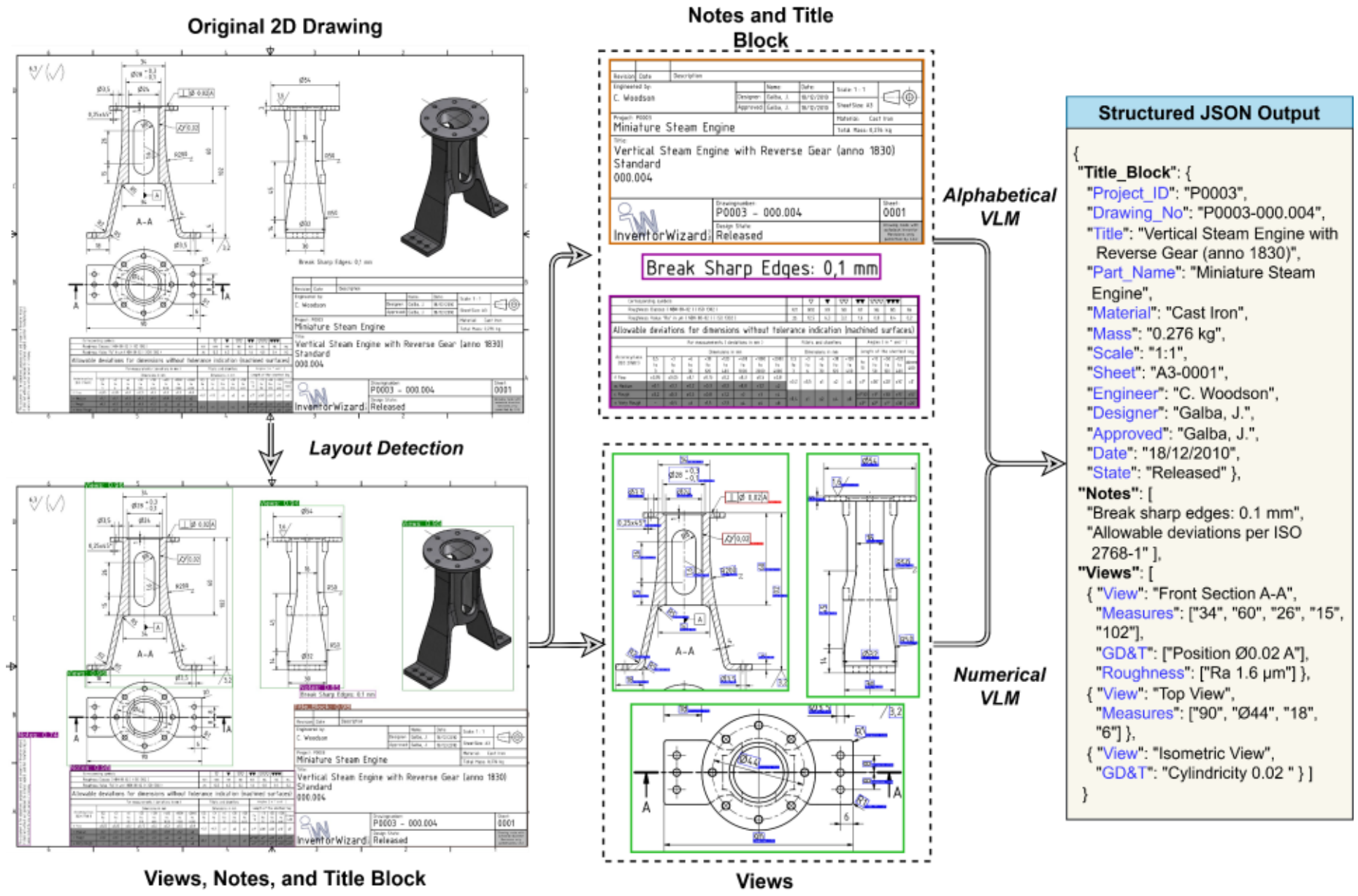

**Fig. 3.** Example of the proposed framework on a sample drawing, illustrating layout detection, fine-grained annotation localization, and semantic parsing to produce a unified JSON output.

### 4.1 Stage 1: Region Detection (Layout Segmentation)

The first stage focuses on identifying the major layout components of a drawing: the drawing *Views*, *Title Block*, and *Notes*. Isolating these high-level regions simplifies downstream processing by providing spatial context and focus for subsequent modules.

This stage was implemented using the YOLOv11-det one-stage object detector, chosen for its high localization accuracy and fast inference speed. The model was trained on the custom annotated layout detection dataset containing 1,000 drawings with three region classes. The dataset was divided into 80 % for training and 20 % for testing to ensure unbiased evaluation. During inference, the detector outputs axis-aligned bounding boxes and class labels for each region type, enabling automatic segmentation of each drawing into logical sub-documents. Each detected *View* is later processed for technical annotations, while *Title Block* and *Notes* regions (typically containing tabular and textual content) are passed to the text-centric parsing stage.

The YOLOv11-det model achieved strong performance in detecting high level regions. As shown in the normalized confusion matrix (Fig. 4), the model achieved 0.96 accuracy for *Views*, 0.99 for *Title Block*, and 0.98 for *Notes*. These results confirm that the detector effectively distinguishes key layout components across diverse drawing styles and template variations.

### 4.2 Stage 2: Detailed View Analysis (Annotation Localization)

The second stage focuses on detecting fine-grained annotations within each localized *View* region. Engineering views often contain numerous rotated or oriented symbols and textual elements, such as dimensions, GD&T feature control frames, and surface roughness marks. To effectively capture these annotations, we employed YOLOv11-obb, a variant of YOLO adapted for OBBs that predicts object angles and anisotropic box dimensions. The model was trained on the 1,406 annotated localization dataset, which includes *Measures*, *GD&T* frames, and *Surface Roughness* instances. The dataset was split into 80 % for training and 20 % for testing. During inference, YOLOv11-obb is applied to each cropped *View* (from Stage 1) to detect and classify fine annotations. This produces a set of OBBs with corresponding labels, which are then passed to the vision language parsing stage.

Processing one view at a time minimizes confusion that might occur if the detector operated on the entire drawing containing overlapping annotations from multiple projections. The oriented box approach ensures robust localization of rotated elements, which traditional axis-aligned detectors or OCR systems often miss. As shown in Fig. 4, the model achieved accuracy of 0.95 for *Measures*, 0.97 for *GD&Ts*, and 0.54 for *Surface Roughness*. The lower accuracy for *Surface Roughness* is primarily due to dataset imbalance, with only 152 instances compared to thousands of *Measures* and *GD&T* samples. The class distribution imbalance reflects real world engineering drawings, where dimensional annotations are far more common than other symbol types. Nevertheless, the model consistently localized and classified annotations across varied orientations, demonstrating strong generalization despite the limited roughness data.

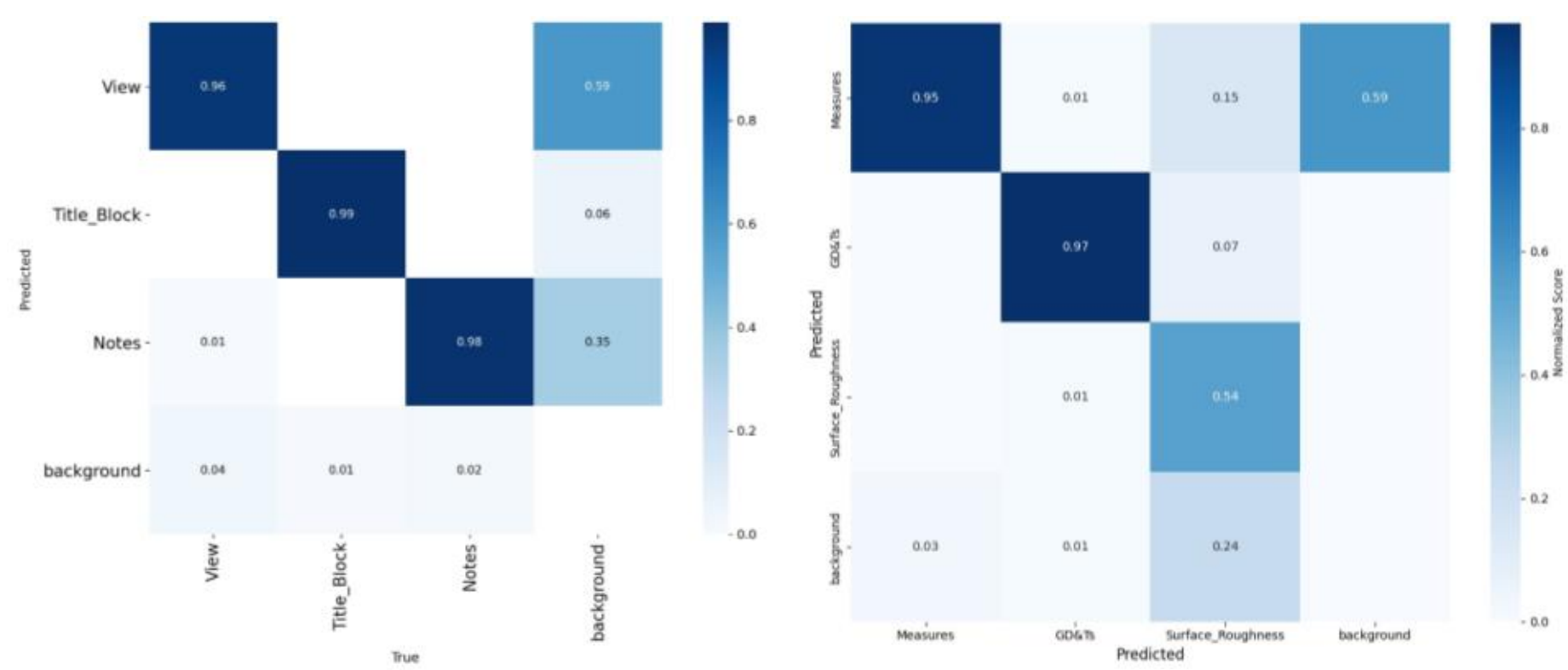

**Fig. 4.** Normalized confusion matrices illustrating detection performance for the two main stages: layout region detection (*left*) and fine-grained annotation localization (*right*).

### 4.3 Stage 3: Vision Language Parsing

Once the relevant regions and annotations are localized, the final stage interprets their semantic content, converting visual data from both text-heavy and numerical regions into a structured digital format.

This stage employs two specialized Donut-based VLMs, selected for their OCR-free, end-to-end multimodal reasoning capability and proven performance in our prior work [8,9]. Two instances of Donut are used: the *Alphabetical* VLM, which processes *Title Block* and *Notes* regions to extract categorical and free-form text; and the *Numerical* VLM, which handles cropped image patches containing *GD&T* symbols, *Measures*, and *Surface Roughness* annotations to extract quantitative specifications. Both models share the same architecture, but only the *Numerical* VLM was fine-tuned on the training dataset, since its output maps to a schema-defined JSON containing Unicode-encoded engineering symbols that provide reliable supervision. In contrast, *Title Block* and *Notes* lack a consistent schema and show high stylistic variability, so the *Alphabetical* VLM was used in zero-shot mode.

The *Numerical* VLM was trained on the dataset described in Section 3.3 using a 70:20:10 split for training, validation, and testing. Fine-tuning was conducted on an NVIDIA GeForce RTX 5090 GPU for 30 epochs, employing the AdamW optimizer with cosine learning rate decay (initial rate of $1\times10^{-6}$, no warm-up). A batch size of four and mixed-precision (FP16) training improved computational efficiency.

Model performance was evaluated using precision, recall, F1 score, and hallucination rate to assess accuracy and reliability. The *Alphabetical* VLM achieved an F1 score of 0.533 for *Title Block* and 0.810 for *Notes*, corresponding to hallucination rates of 0.478 and 0.319, respectively. The overall F1 score of 0.672 indicates that text-heavy and categorical fields remain challenging for OCR-free models. The *Numerical* VLM demonstrated strong and consistent performance across quantitative annotation types, achieving F1 scores of 0.923 for *Measures*, 0.965 for *GD&Ts*, and 1.0 for *Surface Roughness*. These results confirm that the fine-tuned Donut model is highly effective for numerical and symbolic interpretation in engineering drawings. A detailed summary of the quantitative evaluation for both VLMs is presented in Table 1.

After inference, outputs from both VLMs are merged into a unified JSON representation, consolidating *Title Block* fields, *Notes*, and all *Measures* with their associated specifications. This structured output can be directly integrated into CAD, process planning, or manufacturing databases, enabling seamless and automated transfer of engineering information without manual transcription.

**Table 1.** Performance metrics for Alphabetical and Numerical VLMs on the test dataset.

| Category | Precision | Recall | F1 Score | Hallucination |
|---|---|---|---|---|
| ***Alphabetical* VLM** | | | | |
| Title Block | 0.522 | 0.545 | 0.533 | 0.478 |
| Notes | 0.681 | 1.0 | 0.810 | 0.319 |
| Overall | 0.601 | 0.773 | 0.672 | 0.399 |
| ***Numerical* VLM** | | | | |
| Measures | 0.864 | 0.991 | 0.923 | 0.136 |
| GD&Ts | 0.933 | 1.0 | 0.965 | 0.067 |
| Roughness | 1.0 | 1.0 | 1.0 | 0.0 |
| Overall | 0.932 | 0.997 | 0.963 | 0.067 |

## 5 Discussion

The results demonstrate that the proposed method effectively addresses long-standing challenges in the automated interpretation of 2D drawings. By decomposing the overall task into layout detection, annotation localization, and semantic parsing, the system achieves a practical balance between accuracy, interpretability, and computational efficiency. Unlike prior single-stage approaches, this work introduces a hierarchical decomposition that isolates views before interpretation, enabling robust processing of complex multi-view drawings rather than incremental model-level improvements.

The layout segmentation stage provides stable segmentation across diverse templates, ensuring that each functional section of a drawing is processed independently. This structure significantly reduces interference between overlapping or densely annotated regions, a limitation frequently observed in earlier single-pass architectures. The orientation-aware detection further strengthens robustness by recognizing detailed annotations at arbitrary orientations, enabling consistent identification of GD&T frames, measures, and surface finish symbols even under rotation or distortion.

The behavior of both VLMs also offers important insights into domain adaptation. The fine-tuned *Numerical* VLM performed consistently across all quantitative categories, benefiting from a schema-constrained learning setup that minimized output variability and ensured stable predictions. Its ability to preserve symbolic and numeric precision confirms that OCR-free approaches are well suited for high-accuracy technical interpretation. Conversely, the *Alphabetical* VLM exhibited variability when parsing title block tables or low-quality scans. This outcome reflects the inherent complexity of free-form textual fields in 2D engineering documents, where fonts, spacing, and abbreviations differ widely across organizations. Despite these challenges, the model achieved reasonable performance in extracting free-form notes, suggesting that layout-aware pre-training could further improve text-centric understanding.

From an industrial perspective, the framework offers a direct pathway toward digital integration. The unified JSON output enables seamless transfer of extracted data into manufacturing execution systems, CAD/CAM environments, and quality control databases without manual transcription. Unlike conventional OCR pipelines, the proposed approach does not depend on language-specific optical recognition or rule-based post-processing, making it scalable across multilingual and symbol-rich documents. Furthermore, the modular architecture supports continual retraining as new datasets become available, allowing adaptability to evolving drafting conventions and organization-specific templates. Overall, the findings validate that structured, region-aware processing combined with multimodal semantic reasoning provides a practical and extensible foundation for intelligent document understanding in manufacturing.

## 6 Conclusions

This paper presented a three-stage framework for the extraction of structured information from 2D drawings. The architecture integrated YOLOv11-det for layout segmentation, YOLOv11-obb for annotation detection, and Donut-based VLMs for semantic parsing of content. The results confirmed that the system achieved high accuracy in layout and annotation detection and delivered strong performance in numeric data interpretation, offering a reliable and efficient alternative to generic OCR-driven solutions. By producing machine-readable JSON representations, the proposed framework bridges the gap between drawing archives and digital manufacturing ecosystems, enabling automated data flow into CAD/CAM and inspection pipelines.

Despite its strong performance, certain limitations remained. The *Alphabetical* Donut model was not fine-tuned and therefore underperformed for textual data, particularly in title block fields with dense tabular layouts. The *Surface Roughness* class also exhibited lower detection accuracy due to the limited number of annotated training samples. These challenges highlight the need for larger, more balanced datasets and improved handling of text-centric information.

In future work, we plan to integrate advanced VLMs such as GPT-4o and related architectures that excel in text comprehension and structured output generation, aiming to enhance performance on categorical and free-form textual fields. Future research also focuses on synthetic data augmentation for underrepresented annotations and the integration of VLM-based verification modules to cross check generated outputs for consistency. In conclusion, the study demonstrates that combining vision-based layout understanding with OCR-free semantic reasoning constitutes an effective and generalizable solution for automated engineering drawing analysis. The proposed framework represents a robust step toward intelligent, self-improving systems capable of accelerating digital transformation across modern manufacturing industries.

## Acknowledgements

This work is supported by the Singapore International Graduate Award (SINGA) (Awardee: Muhammad Tayyab Khan) funded by A*STAR, an AcRF Tier 1 grant (RG70/23) from Ministry of Education, Singapore, and Nanyang Technological University, and Korea Planning & Evaluation Institute of Industrial Technology (KEIT) grant funded by the Korea government (MOTIE) (No. RS-2024-00442448).